\documentclass[10pt,twocolumn,letterpaper]{article}

\usepackage{cvpr}
\usepackage{times}
\usepackage{epsfig}
\usepackage{graphicx}
\usepackage{amsmath}
\usepackage{amssymb}

\usepackage{bm}
\usepackage{booktabs} 
\usepackage{epstopdf}
\usepackage{url}                                
\usepackage{ulem}
\usepackage{color}
\usepackage{cite}
\usepackage{algorithm,algcompatible,amssymb,amsmath}

\algnewcommand\algorithmicto{\textbf{to}}
\algnewcommand\RETURN{\State \textbf{return} }

\algnewcommand\algorithmicinput{\textbf{Input:}}
\algnewcommand\INPUT{\item[\algorithmicinput]}

\algnewcommand\algorithmicoutput{\textbf{Output:}}
\algnewcommand\OUTPUT{\item[\algorithmicoutput]}

\algnewcommand\algorithmicinitialize{\textbf{Initialize:}}
\algnewcommand\INITIALIZE{\item[\algorithmicinitialize]}
\usepackage[breaklinks=true,bookmarks=false]{hyperref}

\cvprfinalcopy 

\def\cvprPaperID{****} 

\ifcvprfinal\fi

\begin{document}
\title{Revisiting Dilated Convolution: A Simple Approach for Weakly- and Semi- Supervised Semantic Segmentation}


\author{{Yunchao~Wei$^{1}$ \quad Huaxin Xiao$^{2}$ \quad Honghui Shi$^{3}$ \quad Zequn Jie$^{4}$ \quad Jiashi Feng $^{2}$ \quad Thomas S. Huang$^{1}$}\\
	{$^{1}$ UIUC \quad $^{2}$ NUS \quad $^{3}$ IBM Thomas J. Watson Research Center \quad $^{4}$ Tencent AI Lab } \\
	{\tt \small \{wychao1987, huaxinxiao89, shihonghui3, zequn.nus, jshfeng\}@gmail.com \quad huang@ifp.uiuc.edu}
}


\maketitle

\thispagestyle{empty}
\begin{abstract}

Despite the remarkable progress, weakly supervised segmentation approaches are still inferior to their fully supervised counterparts. We obverse the performance gap mainly comes from their limitation on learning to produce high-quality dense object localization maps from image-level supervision. To mitigate such a gap,  we revisit the dilated convolution~\cite{chen2014semantic} and reveal how it can be utilized in a novel way to effectively overcome this critical limitation of weakly supervised segmentation approaches. Specifically, we find that varying dilation rates can effectively enlarge the receptive fields of convolutional kernels and more importantly transfer the surrounding discriminative information to non-discriminative object regions, promoting the emergence of these regions in the object localization maps. Then, we design a generic classification network equipped with convolutional blocks of different dilated rates. It can produce dense and reliable object localization maps and effectively benefit both weakly- and semi- supervised semantic segmentation. Despite the apparent simplicity, our proposed approach obtains superior performance over state-of-the-arts. In particular, it achieves 60.8\% and 67.6\% mIoU scores on Pascal VOC 2012 test set in weakly- (only image-level labels are available) and semi- (1,464 segmentation masks are available) supervised settings, which are the new state-of-the-arts.
\vspace{-3mm}

\end{abstract}
\section{Introduction}

Weakly-supervised image recognition approaches~\cite{pathak2014fully,jie2017deep,zhang2018adversarial,liang2017learning,liang2014computational,yuan2017temporal,zhang2017spftn,zhang2017learning,liu2017surveillance,zhang2017ppr,zhang2017relation} have been extensively studied as they do not require expensive human effort. Among them, the most attractive one is learning to segment images from only image-level annotations. For such approaches, the arguably most critical challenge remaining unsolved is how to accurately and densely localize object regions to obtain high-quality object cues for initiating and improving the segmentation model training~\cite{2015-long,chen2014semantic,zheng2015conditional}.


\begin{figure}[t]
	\centering
	\includegraphics[scale=0.6]{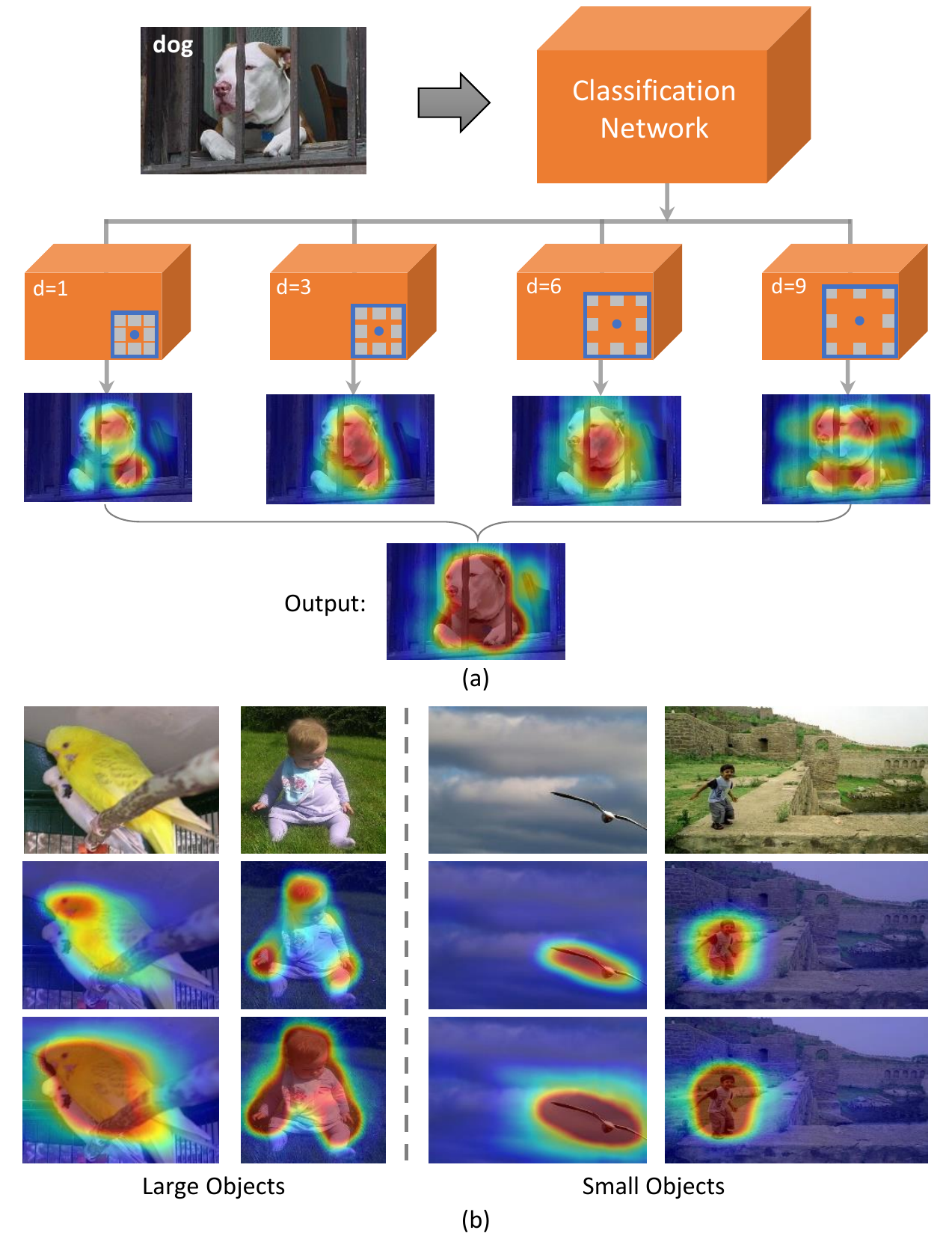}
	\caption{(a) Our proposed approach: equipping a standard classification network with multiple dilated convolutional blocks of different dilation rates for dense object localization. (b) Comparison between the state-of-the-art CAM~\cite{zhou2015cnnlocalization} (the 2nd row) and ours (the last row) on quality of the produced object localization maps. Our approach localizes target objects more accurately even in presence of great scale variation.}
	\label{fig:illu}
	\vspace{-5mm}
\end{figure}
Recently, some top-down approaches~\cite{zhou2015cnnlocalization,zhang2016top} propose to leverage a classification network to produce class-specific attention cues for object localization. However, directly employing attentions produced by image classification models can only identify a small discriminative region of the target object, which is not sufficiently dense and extensive for training a good segmentation model. For instance, some samples of class-specific region localization produced by the state-of-the-art Class Activation Mapping (CAM)~\cite{zhou2015cnnlocalization} are shown in the second row of Figure~\ref{fig:illu} (b). One can observe that CAM hardly generates dense object regions in usual cases where large objects are present, which deviates from requirement of the semantic segmentation task. Those regions discovered by CAM are usually scattered around the target object, \eg some discriminative parts such as head and hands of the \emph{child}. Inability to learn to produce dense object localization from image-level supervision is a critical obstacle to developing well performing weakly supervised segmentation models. Based on such an observation, we propose to transfer discriminative knowledge from those sparsely highlighted regions to adjacent object regions and thus form dense object localization, which can essentially lift segmentation model learning favorably.  



To this end, we revisit the popular dilated convolution and find it indeed provides promising solution up to proper utilization. Dilated convolution was initially introduced by Chen \etal~\cite{chen2014semantic,chen2016deeplab} for semantic segmentation. One key advantage is that it can effectively enlarge receptive field size to incorporate context without introducing extra parameters or computation cost. We find such a feature well fits propagating discriminative information across image regions and highlighting non-discriminative object regions to produce dense object localization. Motivated by this, we introduce multiple dilated convolutional blocks to augment a standard classification model, as shown in Figure~\ref{fig:illu} (a). 

In particular, our proposed approach expands receptive fields at multiple scales by varying dilated rates of convolutional kernels. In general, classification networks are able to identify one or more small discriminative parts with high response for correctly recognizing images. By enlarging the receptive field, object regions with low response can gain improved discriminativeness through perceiving the surrounding high response context. In this way, the discriminative information of high response parts of the target object can propagate to adjacent object regions at multiple scales, making them easier to be identified by classification models. We utilize CAM~\cite{zhou2015cnnlocalization} to generate an object localization map for each convolutional block. As shown in Figure~\ref{fig:illu} (a), the convolution block can only localize two small discriminative regions without enlarging dilation rate, \ie $d=1$. By gradually increasing the dilated rates (from 3 to 9), more object-related regions are discovered. 

However, some true negative regions may be falsely highlighted with large dilated rates (\eg the localization map corresponding to $d=9$). We then propose a simple yet effective anti-noise fusion strategy to address this issue. This strategy can effectively suppress object-irrelevant regions activated by enlarged receptive fields and fuse the localization maps produced by different dilated blocks into an integral one which sharply highlights object regions. From examples shown in Figure~\ref{fig:illu} (b), it can be observed that our approach is very robust to scale variation and is able to densely localize the target objects.

We use the localization maps generated by our proposed approach to produce segmentation masks for training segmentation models. Our approach is generic and can be deployed for learning semantic segmentation networks in both weakly- and semi- supervised manner. Despite its apparent simplicity, our approach indeed provides dense object localization that can easily boost the weakly- and semi- supervised semantic segmentation to new state-of-the-arts, as demonstrated in extensive experiments. To sum up, the main contributions of this work are three-fold:

\begin{itemize}
	\vspace{-2mm}
	\item We revisit the dilated convolution and reveal that it naturally fits the requirement on densely localizing
	object regions for building a good weakly supervised segmentation model, which is new to weakly/semi-supervised image semantic segmentation. 
	\vspace{-1mm}
	\item We propose a simple yet effective approach that leverages dilated convolution to densely localize objects by transferring discriminative segmentation information. 
	\vspace{-1mm}
	\item Our proposed approach is generic for learning semantic segmentation networks in weakly- and semi- supervised manner. In particular, it achieves the mIoU scores of 60.8\% and 67.6\% on test set of Pascal VOC segmentation benchmark in weakly- and semi- settings respectively, which are new state-of-the-arts.

\end{itemize}

\vspace{-3mm}

\section{Related Work}
\vspace{-2mm}
\noindent \textbf{Segmentation with Coarse Annotations} Collecting a large number of pixel-level annotations for training semantic segmentation models is labor intensive. To reduce the burden of pixel-level annotation, Dai \etal~\cite{2015-dai} and Papandreou \etal~\cite{2015-papandreou-weakly} proposed to learn semantic segmentation with annotated bounding boxes. Lin \etal~\cite{lin2016scribblesup} employed semantic scribbles as supervision for semantic segmentation. More recently, the supervised annotation is further relaxed to instance points in~\cite{russakovsky2015s}.

\noindent \textbf{Segmentation with Image-level Annotations} Image-level label, which is easy to obtain, is the simplest supervision for leaning to segment. Some works~\cite{pinheiro2015weakly,pathak2015constrained,pathak2014fully} proposed to utilize multiple instance learning for semantic segmentation with image-level labels. Papandreou \etal~\cite{2015-papandreou-weakly} proposed to dynamically predict foreground objects and background for supervision based on an Expectation-Maximization algorithm. Recently, great progress~\cite{kolesnikov2016seed,saleh2016built,qi2016augmented,shimoda2016distinct,wei2015stc,wei2016learning,hong2017weakly,kim2017two,hou2016bottom} has been made on this challenging task. Wei \etal~\cite{wei2016learning} and Qi \etal~\cite{qi2016augmented} utilized proposals to generate pixel-level annotations for supervision. However, making use of MCG~\cite{pont2015multiscale} proposals or adopting additional network~\cite{wei2015hcp} for proposal-based classification usually leads to large time consumption and stronger supervision is also inherently used (MCG has been trained from PASCAL \emph{train} images with pixel-level annotations). Wei \etal~\cite{wei2015stc} presented a simple to complex (STC) framework to progressively improve the ability of the segmentation network. However, the success of STC mainly depends on a large number of simple images for training. Kolesnikov \etal~\cite{kolesnikov2016seed} proposed an SEC approach that integrates three loss functions, \ie seeding, expansion and constrain-to-boundary, into a unified framework to train the segmentation network. But SEC can only obtain small and sparse object-related seeds for supervision, which can not provide enough information for leaning reliable segmentation models. Most recently, Wei \etal~\cite{wei2017object} proposed an adversarial erasing (AE) approach to mine dense object regions for supervision. Although it achieves the state-of-the-art performance on the PASCAL VOC benchmark, the AE method requires repetitive training procedures to learn multiple classification models, which are then applied to locate object-related regions. Comparatively, we only need to train one classification model for localizing dense and integral object regions in this work.

\vspace{-2mm}
\section{The Proposed Approach}

\begin{figure}[t]
	\centering
	\includegraphics[scale=0.75]{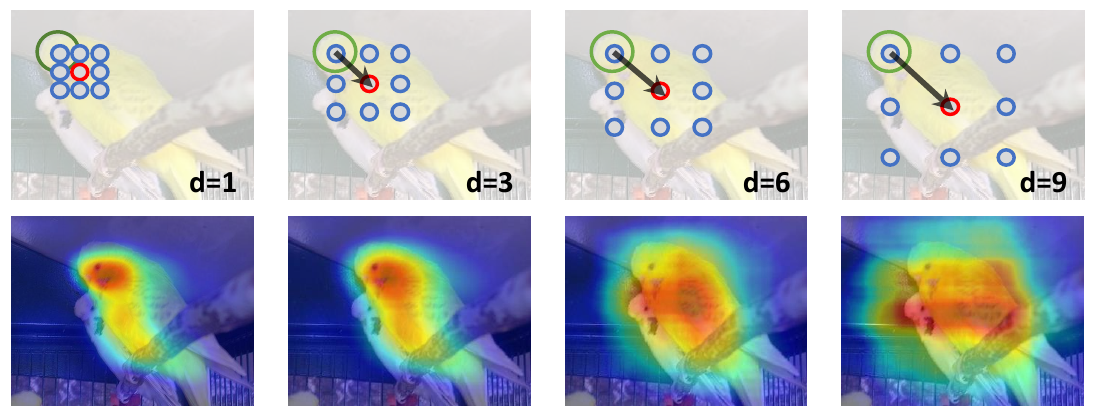}
	\caption{Motivation of our approach: information can be transferred from the initially discriminative region to other regions by varying dilated rates of convolutional kernels. The corresponding localization maps are shown in the 2nd row. Best viewed in color.}
	\label{fig:re-visit}
	\vspace{-4.5mm}
\end{figure}

\subsection{Revisiting Dilated Convolution}
Some top-down approaches~\cite{zhou2015cnnlocalization,zhang2016top} can identify the discriminative object regions contributing to a classification network decision but they generally miss non-discriminative object regions. We propose to augment the classification model by enabling the information to transfer from discriminative regions to adjacent non-discriminative regions to overcome such a limitation. We find that dilated convolution~\cite{chen2014semantic}, which can effectively incorporate surrounding context by enlarging receptive field size of kernels, provides a promising solution. Figure~\ref{fig:re-visit} illustrates how dilation enables information transfer. Originally, the \emph{head} region in the green cycle is most discriminative for the classification network to recognize this as a ``bird" image. We adopt a 3x3 convolutional kernel to learn the following feature representation at the location indicated by the red cycle. By enlarging the dilated rate from 1 to 3 of a 3$\times$3 kernel, the location near the \emph{head} will be perceived and get their discriminativeness enhanced. By further increasing the dilated rates (to $d=6, 9$), some further locations will perceive the \emph{head} and similarly facilitate the classification model to discover these regions.
To prove the dilated convolution can indeed improve the discriminative ability of low response object regions, we produce the localization maps at different dilated rates using CAM~\cite{zhou2015cnnlocalization}. We can observe that those low response object regions on the localization map of $d=1$ can be effectively highlighted with various dilated rates. The produced localization maps are complementary according to different dilated rates, and thus integrating results from multiple dilated blocks is also necessary.

\begin{figure}[t]
	\centering
	\includegraphics[scale=0.72]{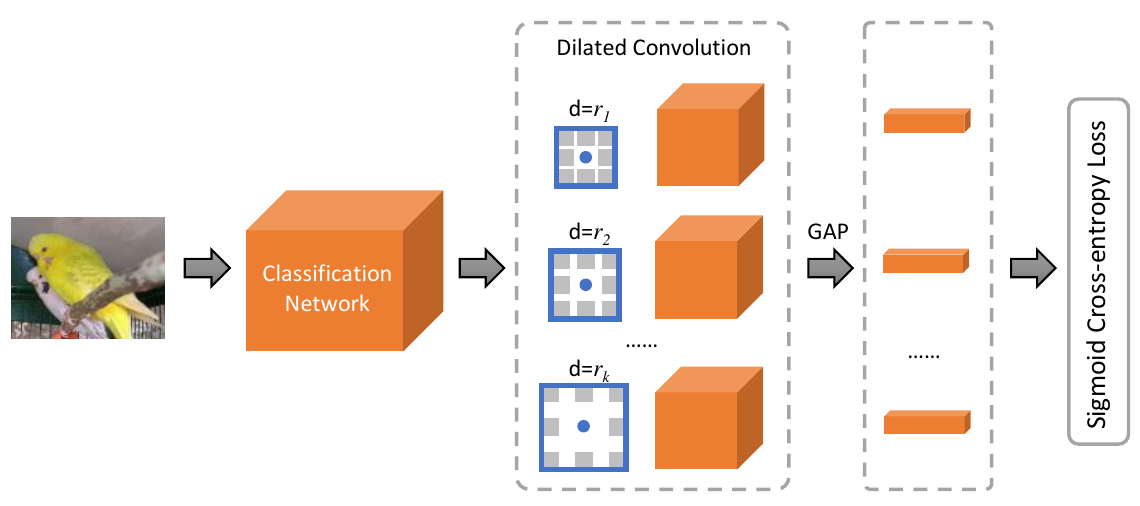}
	\caption{Illustration on training the network with multiple dilated convolutional blocks.}
	\label{fig:cls_train}
	\vspace{-3mm}
\end{figure}

\subsection{Multi-dilated Convolution for Localization}
Motivated by the above findings, we present an augmented classification network with multi-dilated convolutional (MDC) blocks to produce dense object localization, as shown in Figure~\ref{fig:cls_train}. The network is built upon the VGG16~\cite{simonyan2014very} model pre-trained on ImageNet~\cite{2009-imagenet}. We remove those fully-connected layers, and one pooling layer to enlarge the resolution of feature maps. Then, convoluational blocks with multiple dilated rates (\ie $d=r_i$, $i=1,\cdots, k$) are appended to \emph{conv5} to localize object-related regions perceived by different receptive fields. After global average pooling (GAP), the produced representations further pass through a fully-connected layer to predict image-level classification. We optimize the classification network by minimizing sigmoid cross-entropy loss, and the classification activation maps (CAM)~\cite{zhou2015cnnlocalization} approach is then employed to produce the class-specific localization map for each block.

We implement two kinds of convolutional operations. 1) We apply the standard kernels, \ie $d=1$. In this case, we can obtain accurate localization maps in which some discriminative parts of the target object are highlighted but many object-related regions are missed. 2) To transfer the discriminative knowledge of sparsely highlighted regions to other object regions, we vary dilated rates to enlarge the receptive field of kernels. In this way, the discriminative features from the adjacent highlighted regions can be transferred to the object-related regions that have not been discovered. We observe that convolutional blocks of large dilated rates will introduce some irrelevant regions, \ie some true negative regions highlighted by taking advantage of adjacent discriminative object parts. Therefore, we propose to use small dilation rates (\ie, $d=3,6,9$) in this work. 

However, a few unrelated regions may still be identified even if we adopt small dilation rates. To address this issue, we propose a simple anti-noise fusion strategy to suppress object-irrelevant regions and fuse the generated localization maps into an integral one where the object regions are sharply highlighted. We note that the true positive object-related regions are usually distinguishable by two or more localization maps and the true negative regions show diversity under different dilations. To anneal the false regions, we conduct an average operation over the localization maps generated by different dilated convolutional blocks ($d=3,6,9$). Then, the averaged map is added to the localization map of the standard convolutional block ($d=1$) to produce the final localization map. In this way, the accurate regions mined by standard convolutional blocks are not missed. Formally, we use $H_0$ and $H_i$ ($i=1\cdots n_d$ and $n_d$ is the number of dilated convolutional blocks) to denote the localization maps generated by standard and dilated convolutional blocks, respectively. The final localization map $H$ for object region generation is then produced by $H=H_0 + \frac{1}{n_d}\sum_{i=1}^{n_d}{H_i}.$

\begin{figure}[t]
	\centering
	\includegraphics[scale=0.45]{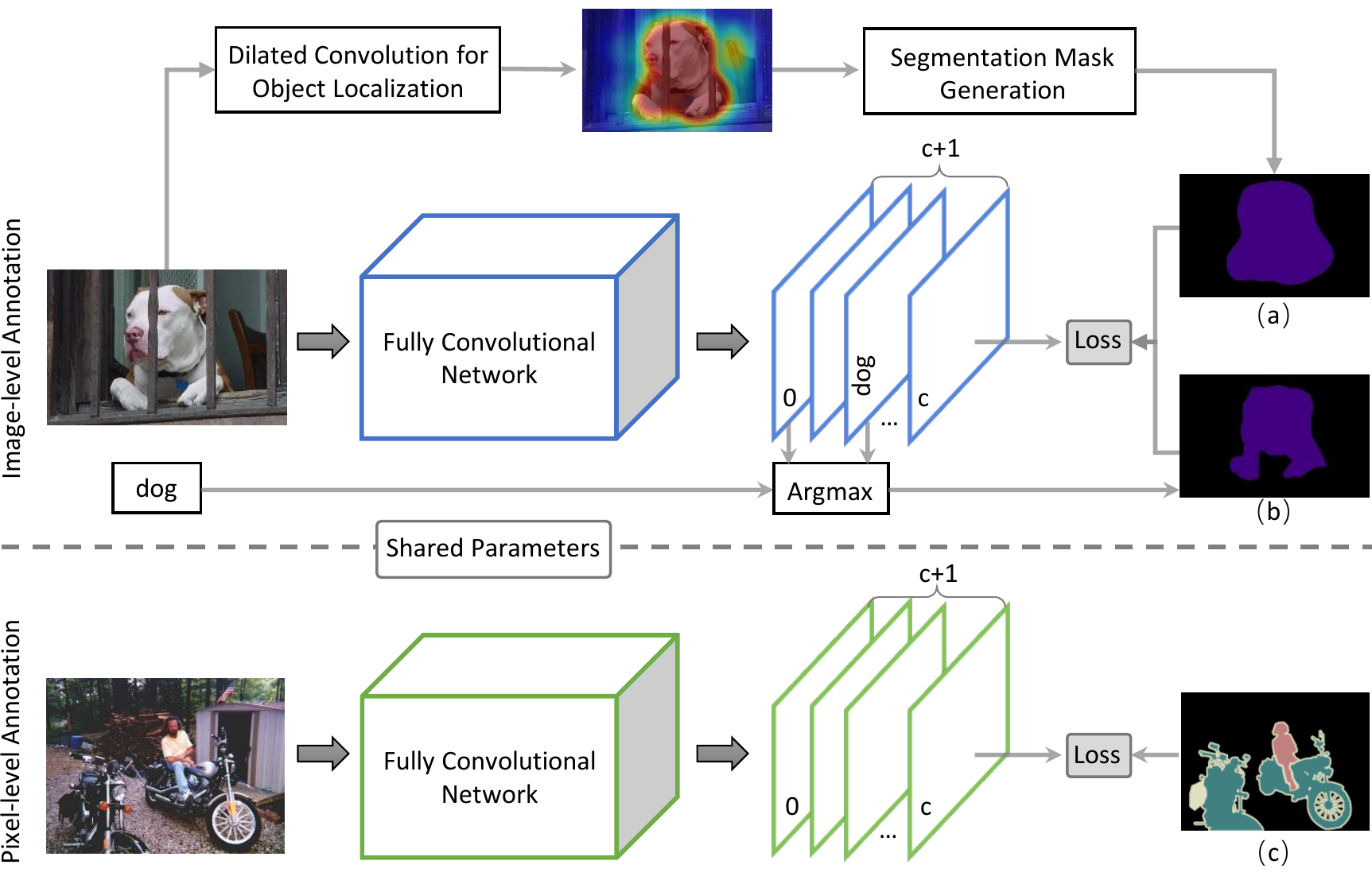}
	\caption{Details of training semantic segmentation in weakly- or semi- supervised manner with our proposed approach. In particular, (a) is the segmentation mask inferred from the dense localization map; (b) is the online predicted segmentation mask; (c) is the human annotated segmentation mask.}
	\label{fig:framework}
	\vspace{-4mm}
\end{figure}
Based on $H$, the pixels with values larger than a pre-defined threshold $\delta$ are considered as foreground supportive object-related regions. Besides, background localization cues are also needed for training the segmentation network. Motivated by~\cite{kolesnikov2016seed,wei2017object,wei2015stc} , we utilize the saliency detection method~\cite{xiao2017self} to produce saliency maps for training images and take the pixels with low saliency values as background. We follow the same strategy detailed in~\cite{wei2017object} to merge the highlighted object regions and the background cues. Finally, we are able to obtain the predicted segmentation mask of each training image for learning to segment. 

\subsection{Weakly- and Semi- Segmentation Learning}
We apply the dense localization maps produced by the proposed approach for training weakly and semi- supervised segmentation models. 

\vspace{-3mm}
\subsubsection{Weakly-supervised Learning}
For the weakly-supervised application, we adopt a similar framework as the one proposed in~\cite{2015-papandreou-weakly, wei2017object} to exploit those ignored pixels in the inferred segmentation masks and gain robustness to falsely labeled pixels, as shown in the upper part of Figure~\ref{fig:framework}. In particular, we extract the confidence maps corresponding to ground truth image-level labels for inferring segmentation masks in an online manner, which together with the segmentation masks derived from dense localization maps serve as supervision.

We explain the process more formally. Let $I_w$ denote an image from the weakly-supervised training set $\mathcal{I}_w$. For any $I_w \in \mathcal{I}_w$, $M_w$ is the corresponding pseudo segmentation mask produced by a dense localization map and $\mathcal{C}$ is the label set where \emph{background} category is also included. Our target is to train a segmentation model (\eg FCN) $f(I_w;\theta)$ with learnable parameter $\theta$. The FCN models the conditional probability of any label $c \in \mathcal{C}$ at any location $u$ of the class-specific confidence map $f_{u,c}(I_w; \theta)$. Use $\hat{M}_w$ to denote the online predicted segmentation mask of $I_w$, which collaborates with $M_w$ for supervision. The loss function for optimizing the weakly-supervised FCN is formulated as
\begin{equation}
\min\limits_{\theta}\sum\limits_{I_w \in \mathcal{I}_w}{J_{w}(f(I_w;\theta))},\vspace{-0.5em}
\end{equation}
where
\vspace{-0.5em}
\[
\begin{aligned}
J_{w}(f(I_w;\theta)) = &-{\frac{1}{\sum\limits_{c \in \mathcal{C}}|M_w^c|} }\sum\limits_{c \in \mathcal{C}}\sum\limits_{u \in M_w^c}\log f_{u,c}(I_w;\theta)\\&-{\frac{1}{\sum\limits_{c \in \mathcal{C}}|\hat{M}_w^c|} }\sum\limits_{c \in \mathcal{C}}\sum\limits_{u \in \hat{M}_w^c}\log f_{u,c}(I_w;\theta),\vspace{-0.5em}
\end{aligned}
\]
and $|\cdot|$ indicates the number of pixels.
\vspace{-2.5mm}
\subsubsection{Semi-supervised Learning}
Along with a large quantity of images with image-level annotations, we are interested in utilizing pixel-level annotations over a small number of images to further push the segmentation performance, \ie the semi-supervised learning setting. As shown in the bottom of Figure~\ref{fig:framework}, both strongly and weakly annotated images can be easily combined to learn segmentation networks by sharing parameters.

Let $I_s$ denote an image from the strongly-supervised training set $\mathcal{I}_s$ and $M_s$ is the corresponding segmentation mask annotated by human. The loss function used for optimizing the semi-supervised FCN can be defined as
\begin{equation}
\label{eq:ws}
\min\limits_{\theta}\sum\limits_{I_w \in \mathcal{I}_w}{J_{w}(f(I_w;\theta))} + \sum\limits_{I_s \in \mathcal{I}_s}{J_{s}(f(I_s;\theta))},\vspace{-0.5em}
\end{equation}
where
\[
J_{s}(f(I_w;\theta)) = -{\frac{1}{\sum\limits_{c \in \mathcal{C}}|M_s^c|} }\sum\limits_{c \in \mathcal{C}}\sum\limits_{u \in M_s^c}\log f_{u,c}(I_s;\theta).
\]


\section{Experiments}
\subsection{Dataset and Settings}
\noindent \textbf{Dataset and Evaluation Metrics} The proposed approach is evaluated on the PASCAL VOC 2012 segmentation benchmark~\cite{2010-pascal}. One background category and 20 object categories are annotated in this dataset. Following the common practice~\cite{chen2014semantic,hariharan2011semantic,wei2017object}, the number of training images is increased to 10,582 by augmentation. The validation and test subsets include 1,449 and 1,456 images, respectively. We evaluate the performance in terms of pixel mIoU averaged on 21 categories. For all experiments, only image-level labels are employed as supervision and detailed analysis is conducted on the validation set. We compare our approach with other state-of-the-arts on both validation and test sets. Those results on the test set are obtained by submitting the predicted results to the official PASCAL VOC evaluation server.

\noindent \textbf{Training/Testing Setting} We adopt the convolutional layers of VGG16~\cite{simonyan2014very} pre-trained on ImageNet~\cite{2009-imagenet} to initialize the classification network except for the new added convolutional blocks. For the segmentation network, the DeepLab-CRF-LargeFOV model from~\cite{chen2014semantic} is selected as the basic network, whose parameters are also initialized by VGG16. We take a mini-batch size of 30 images. Patches of 321$\times$321 pixels are randomly cropped from images for training both classification and segmentation networks. We train the model for 15 epochs. The initial learning rate is set to 0.001 and decreased by a factor of 10 after 6 epochs. All the experiments are performed on NVIDIA TITAN X PASCAL GPU. We use the DeepLab~\cite{chen2014semantic} code, which is implemented based on the publicly available Caffe framework~\cite{jia2014caffe}. To obtain the object-related region based on the dense localization map, the pixels belonging to the top 30\% of the unique largest value are selected as object regions. Saliency maps produced by~\cite{xiao2017self} are utilized to provide background cues. Following the settings of~\cite{wei2017object}, we set the pixels with normalized saliency values smaller than 0.06 as background. All the conflicted and unassigned pixels are ignored for training.


\begin{table}[]
	\centering
	\caption{Comparison of weakly-supervised semantic segmentation methods on PASCAL VOC 2012 validation and test sets.}
	\small
	\label{tab:val-comp}
	\begin{tabular}{lccc}
		\toprule
		Methods & Training Set & validation & test \\
		\midrule
		\multicolumn{2}{l}{Supervision: Scribbles}  \\
		Scribblesup {\tiny{CVPR2016}}~\cite{lin2016scribblesup} & 10K & 63.1 & -\\
		\midrule
		\multicolumn{2}{l}{Supervision: Box}  \\
		WSSL {\tiny{ICCV2015}}~\cite{2015-papandreou-weakly} & 10K & 60.6 & 62.2\\
		BoxSup {\tiny{ICCV2015}}~\cite{2015-dai}  & 10K & 62.0 & 64.2\\		
		\midrule
		\multicolumn{2}{l}{Supervision: Spot}  \\
		1 Point {\tiny{ECCV2016}}~\cite{russakovsky2015s} & 10K & 46.1 & -\\
		Scribblesup {\tiny{CVPR2016}}~\cite{lin2016scribblesup} & 10K & 51.6 & -\\		
		\midrule
		\multicolumn{2}{l}{Supervision: Image-level Labels}  \\			
		MIL-FCN {\tiny{ICLR2015}}~\cite{pathak2014fully} & 10K & 25.7 & 24.9\\
		CCNN {\tiny{ICCV2015}}~\cite{pathak2015constrained} & 10K & 35.3 & 35.6\\
		EM-Adapt {\tiny{ICCV2015}}~\cite{2015-papandreou-weakly} & 10K & 38.2 & 39.6\\
		MIL-seg* {\tiny{CVPR2015}}~\cite{pinheiro2015weakly} & 700K & 42.0 & 40.6\\
		SN\_B* {\tiny{PR2016}}~\cite{wei2016learning} & 10K & 41.9 & 43.2\\		
		TransferNet* {\tiny{CVPR2016}}~\cite{hong2015learning} & 70K & 52.1 & 51.2\\								
		DCSM {\tiny{ECCV2016}}~\cite{shimoda2016distinct} & 10K & 44.1 & 45.1\\
		BFBP {\tiny{ECCV2016}}~\cite{saleh2016built} & 10K & 46.6 & 48.0\\
		SEC {\tiny{ECCV2016}}~\cite{kolesnikov2016seed} & 10K & 50.7 & 51.7\\				
		AF-MCG* {\tiny{ECCV2016}}~\cite{qi2016augmented} & 10K &  54.3 & 55.5\\
		STC {\tiny{TPAMI2017}}~\cite{wei2015stc} & 50K & 49.8 & 51.2\\
		Saleh \etal{\tiny{TPAMI2017}}~\cite{saleh2017incorporating} & 10K & 50.9 & 52.6 \\ 
		Ray \etal {\tiny{CVPR2017}}~\cite{roy2017combining} & 10K & 52.8 & 53.7 \\
		AE-PSL {\tiny{CVPR2017}}~\cite{wei2017object} & 10K & 55.0 & 55.7\\	
		Hong \etal {\tiny{CVPR2017}}~\cite{hong2017weakly} & 970K & 58.1 & 58.7 \\
		Kim \etal {\tiny{ICCV2017}}~\cite{kim2017two} & 10K & 53.1 & 53.8 \\
		MDC (Ours) & 10K & $\bm{60.4}$ & $\bm{60.8}$\\
		\bottomrule	
		\multicolumn{4}{l}{(* indicates methods implicitly use pixel-level supervision)}\\
	\end{tabular}	
	\vspace{-4mm}
\end{table}

\subsection{Comparison with State-of-the-arts}
\subsubsection{Weakly-supervised Semantic Segmentation}
For weakly-supervised semantic segmentation, we mainly compare the approaches using coarse pixel-level annotation (including scribbles, bounding boxes and spots) and image-level annotation as supervision information. Table~\ref{tab:val-comp} displays the comparison on the PASCAL VOC validation set and test set. Note that some approaches utilize more images for training, \ie MIL-*~\cite{pinheiro2015weakly} (700K), TransferNet~\cite{hong2015learning} (70K), STC~\cite{wei2015stc} (50K) and Hong \etal~\cite{hong2017weakly} (970K). In addition, pixel-level supervision is implicitly used by some approaches (\eg SN\_B~\cite{wei2016learning} and AF-MCG~\cite{qi2016augmented}) due to using MCG~\cite{pont2015multiscale} proposals. 

From Table~\ref{tab:val-comp}, it can be observed that the segmentation masks inferred from our produced dense localization maps are very reliable for learning segmentation networks, which outperforms all the other approaches using image-level labels as weak supervision. We note that Hong \etal~\cite{hong2017weakly} achieved the state-of-the-art performance on this challenging task. However, the improvement mainly benefits from using additional video data for training. Since temporal dynamics in videos can provide rich information, it is more easily to distinguish the entire object regions from videos than that from still images. Notably, we only use 10K images for training the model which outperforms Hong \etal~\cite{hong2017weakly} by 2.3\% on the validation set. This well demonstrates the effectiveness of the proposed approach on generating high-quality dense object localization maps. AE-PSL needs to conduct multiple adversarial erasing steps to mine object-related regions, which requires training multiple different classification models for object localization. The proposed approach only needs to train one single classification model for localizing object regions and achieves much better mIoU scores than AE-PSL. Compared with AF-MCG~\cite{qi2016augmented}, our approach does not require a huge number of proposals, and thus is more efficient as producing proposals and training on them are time consuming. Without any pixel-level supervision, our weakly-supervised results further approach those of scribble-based and box-based methods and outperform the spot-based approaches by more than 8.8\%. We conduct additional comparison on PASCAL VOC testing set. Our method achieves the new state-of-the-art on this competitive benchmark, and outperforms the mIoU scores of others by more than 2.1\%.

\begin{table}[]\setlength{\tabcolsep}{5pt}
	\centering
	\caption{Comparison of semi-supervised semantic segmentation methods on PASCAL VOC 2012 validation and test sets.}
	\small
	\label{tab:semi-comp}
	\begin{tabular}{lcc}
		\toprule
		Methods & validation & test \\
		
		\midrule
		Weakly Supervision: Boxes & & \\
		BoxSup {\tiny{ICCV2015}}~\cite{2015-papandreou-weakly}  & 63.5 & 66.2 \\
		WSSL {\tiny{ICCV2015}}~\cite{2015-papandreou-weakly}  & 65.1 & 66.6 \\
		Khoreva	\etal {\tiny{CVPR2017}}~\cite{khoreva2017simple}  & 65.8 & 66.9 \\
		\midrule
		Weakly Supervision: Image-level Labels & & \\
		WSSL {\tiny{ICCV2015}}~\cite{2015-papandreou-weakly}  & 64.6 & 66.2\\
		MDC (Ours)  & $\bm{65.7}$ & $\bm{67.6}$\\
		\bottomrule	
	\end{tabular}
\vspace{-2mm}	
\end{table}
\vspace{-3mm}
\subsubsection{Semi-supervised Semantic Segmentation}
For semi-supervised semantic segmentation, we mainly compare with WSSL~\cite{2015-papandreou-weakly} whose weakly annotations are image-level labels. To further validate the quality of dense localization maps, we also compare with approaches that have access to bounding boxes for supervision. We adopt the same strong/weak split as those baselines, \ie 1.4K strongly annotated images and 9K weakly annotated images.

From Table~\ref{tab:semi-comp}, our approach achieves better results than WSSL under the same setting, \ie 65.7\% \emph{vs.} 64.6\% on the validation set and 67.6\% \emph{vs.} 66.2\% on the test set. Furthermore, we also compare with other approaches which use object bounding boxes as weakly-supervised information instead of image-level labels. Even though our approach uses much weaker supervision, it still achieves competitive and better mIoU scores on validation and test sets, respectively.

\begin{figure*}[t]
	\centering
	\includegraphics[scale=0.60]{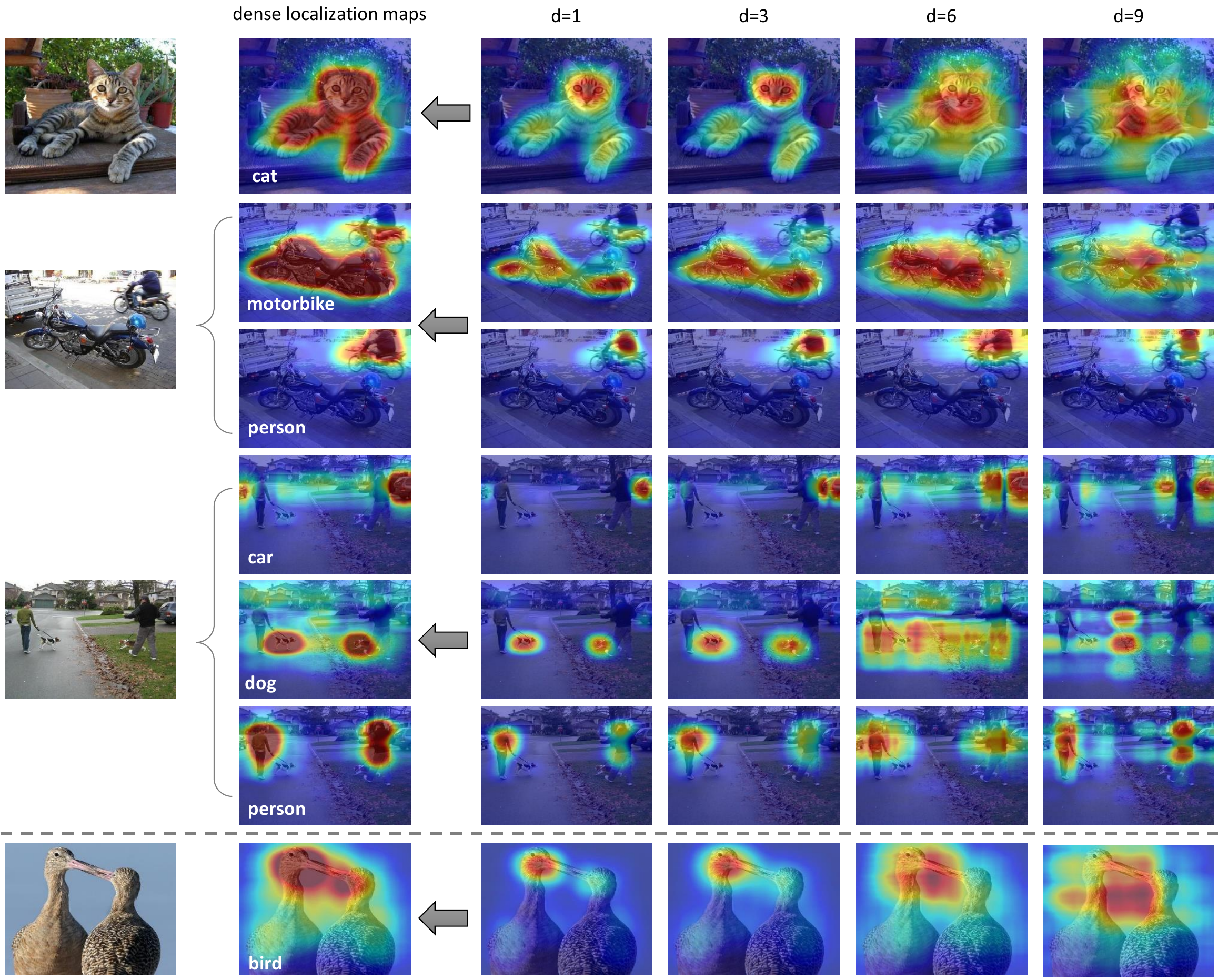}
	\caption{Examples of the localization maps produced by different dilated blocks as well as the dense localization maps with the anti-noise fusion strategy. One failure case is shown at the bottom row.}
	\label{fig:visualize}
	\vspace{-2mm}
\end{figure*}

\begin{table*}[htb]\setlength{\tabcolsep}{2.4pt}
	\centering
	\caption{Comparison of mIoU scores using different localization maps on PASCAL VOC 2012.}
	\label{tab:ablation_weak}
	\footnotesize
	\begin{tabular}{lccccccccccccccccccccc|c}
		\toprule
		settings  & bkg & plane & bike  & bird  & boat  & bottle & bus   & car   & cat   & chair & cow   & table & dog   & horse & motor & person & plant & sheep & sofa  & train & tv &  mIoU  \\
		\midrule
		\multicolumn{22}{l}{Results on the validation set:}\\
		d=1 & 87.0  & 76.1  & 31.4  & 67.7  & 54.9  & 58.0  & 24.9  & 55.1  & 73.7  & 2.6   & 62.6  & 0.3   & 70.3  & 61.8  & 65.0  & 67.5  & 15.8  & 68.2  & 15.1  & 68.0  & 29.6  & 50.3 \\
		d=3 & 87.2  & 75.8  & 31.7  & 66.9  & 54.0  & 58.1  & 33.6  & 57.9  & 73.4  & 5.2   & 61.9  & 1.7   & 70.0  & 62.3  & 65.7  & 67.3  & 18.5  & 68.2  & 16.9  & 68.8  & 32.9  & 51.3 \\
		d=6 & 87.8  & 77.0  & 32.3  & 67.1  & 55.6  & 59.5  & 48.0  & 62.6  & 73.6  & 9.5   & 62.5  & 6.3   & 69.4  & 60.4  & 66.0  & 66.1  & 28.6  & 68.2  & 21.2  & 69.7  & 41.8  & 54.0 \\
		d=9 & 87.9  & 76.5  & 32.1  & 68.0  & 56.1  & 59.2  & 51.3  & 62.9  & 73.0  & 9.3   & 63.7  & 6.2   & 68.0  & 60.7  & 66.0  & 65.0  & 31.0  & 69.3  & 22.9  & 69.3  & 44.1  & 54.4 \\
		fusion & 88.5  & 77.9  & 32.5  & 68.3  & 56.7  & 59.9  & 64.2  & 70.6  & 73.2  & 17.0  & 63.7  & 12.2  & 69.8  & 62.7  & 67.5  & 68.5  & 32.9  & 68.1  & 24.8  & 70.3  & 49.5  & 57.1 \\
		fusion (CRF) & 89.5  & 85.6  & 34.6  & 75.8  & 61.9  & 65.8  & 67.1  & 73.3  & 80.2  & 15.1  & 69.9  & 8.1   & 75.0  & 68.4  & 70.9  & 71.5  & 32.6  & 74.9  & 24.8  & 73.2  & 50.8  & 60.4 \\
		\midrule
		\multicolumn{22}{l}{Results on the test set:}\\
		fusion (CRF) & 89.8  & 78.4  & 36.2  & 82.1  & 52.4  & 61.7  & 64.2  & 73.5  & 78.4  & 14.7  & 70.3  & 11.9  & 75.3  & 74.2  & 81.0  & 72.6  & 38.8  & 76.7  & 24.6  & 70.7  & 50.3  & 60.8 \\
		\bottomrule
	\end{tabular}
\vspace{-4mm}
\end{table*}

\begin{table*}[htb]\setlength{\tabcolsep}{2pt}
	\centering
	\caption{Comparison of mIoU scores using different strong/weak splits on PASCAL VOC 2012.}
	\label{tab:ablation_semi}
	\footnotesize
	\begin{tabular}{lccccccccccccccccccccc|c}
		\toprule
		settings  & bkg & plane & bike  & bird  & boat  & bottle & bus   & car   & cat   & chair & cow   & table & dog   & horse & motor & person & plant & sheep & sofa  & train & tv &  mIoU  \\
		\midrule
		\multicolumn{22}{l}{Results on the validation set:}\\
		strong 500 & 90.4  & 78.3  & 39.4  & 71.6  & 59.9  & 56.0  & 79.8  & 75.0  & 70.0  & 28.3  & 63.7  & 40.6  & 62.1  & 65.2  & 69.3  & 72.0  & 38.4  & 69.8  & 37.6  & 71.9  & 59.3  & 61.8 \\
		strong 1K & 90.6  & 77.9  & 38.6  & 71.2  & 61.5  & 56.4  & 80.1  & 74.7  & 71.0  & 27.3  & 65.1  & 39.8  & 64.9  & 63.8  & 69.8  & 71.8  & 38.3  & 72.1  & 37.1  & 72.7  & 60.7  & 62.2 \\
		strong 1.4K & 90.6  & 78.7  & 40.0  & 73.2  & 62.2  & 56.3  & 80.6  & 75.5  & 70.9  & 26.7  & 66.8  & 42.1  & 64.4  & 65.3  & 69.2  & 72.8  & 38.6  & 72.4  & 37.5  & 73.7  & 59.5  & 62.7 \\
		strong 1.4K (CRF) & 91.7  & 83.8  & 41.5  & 78.3  & 63.7  & 61.8  & 83.9  & 77.6  & 75.8  & 28.6  & 73.8  & 42.9  & 68.5  & 73.2  & 72.7  & 75.7  & 34.2  & 79.1  & 38.5  & 74.1  & 60.7  & 65.7 \\
		\midrule
		\multicolumn{22}{l}{Results on the test set:}\\
		strong 1.4K (CRF) & 92.3  & 82.1  & 46.1  & 76.8  & 55.3  & 58.7  & 83.4  & 78.5  & 79.4  & 27.1  & 74.5  & 50.6  & 73.0  & 76.1  & 83.1  & 76.1  & 48.0  & 81.7  & 44.9  & 73.1  & 59.5  & 67.6 \\
		strong 2.9K (CRF) & 92.4  & 81.1  & 43.6  & 84.0  & 54.5  & 61.0  & 83.3  & 78.7  & 81.5  & 26.1  & 71.2  & 55.5  & 75.4  & 77.3  & 82.2  & 77.1  & 54.3  & 80.3  & 45.8  & 74.0  & 59.4  & 68.5 \\
		\bottomrule
	\end{tabular}
\vspace{-4mm}
\end{table*}

\vspace{-1mm}
\subsection{Ablation Analysis}
\vspace{-1mm}
We then analyze the effectiveness of the proposed dense object localization approach, and how it benefits both weakly- and semi- supervised semantic segmentation.

\vspace{-3mm}
\subsubsection{Strategy of Dense Object Localization}

The adopted classification network for object localization is augmented with convolutional blocks with multiple dilation rates. The object-related cues from different dilated blocks can be integrated into dense and integral object regions. To verify this, samples of localization maps from different convolutional blocks and the fused results are visualized in Figure~\ref{fig:visualize}. We observe that the block ($d=1$) is able to localize objects with high precision but low recall (most regions of the target object are missed). By making use of other blocks with larger dilations ($d=3, 6, 9$), some other object-related regions are highlighted, \eg the \emph{body} of the right \emph{cat} ($d=6$) in the first row and some parts of the \emph{motorbike} in the second row ($d=3$ and $d=6$). However, we note that some true negative regions are also highlighted if we adopt large dilation rates (\eg those localization maps corresponding to $d=6$ and $d=9$). For instance, we can observe that the center region at the map (row 5, column 6) becomes discriminative for the category \emph{dog}. The reason is that the enlarged kernel perceives the context around two dogs when convolutional operation is conducted for the center pixels, which improves the discriminative ability of the produced convolutional features. 

It can be observed that the true positive object-related regions are usually shared by two or more localization maps and the false positive regions are different according to dilation rates. To prevent the false object-related regions from being highlighted, we make an average operation of these localization maps with enlarged dilation rates. Then, we sum the obtained localization map with that produced by the block of $d=1$ to generate the final result. From Figure~\ref{fig:visualize}, we can see that most of the regions of objects are highlighted in the final fused localization maps even for some challenging cases such as multi-class and multi-instance.  

In addition, one merit of our approach worthy of being highlighted is that we can easily use a fixed threshold to obtain most of the object regions accurately based on the generated dense localization map, regardless of the scale of the object. However, it is very difficult to use a fixed threshold to accurately extract the object regions for the localization maps without enlarging dilation rate (\ie $d=1$) as shown in Figure~\ref{fig:illu} (b) and Figure~\ref{fig:visualize}. In particular, we need a small threshold for the large objects so that most object-related regions are discovered. Nevertheless, the value needs to be large for the small objects so that true negative regions can be prohibited. 

We demonstrate one failure case at the bottom row of Figure~\ref{fig:visualize}. This sample is with the following characteristics, \ie the object with large scale and the discriminative regions only sparsely highlighted at one end of the target object when $d=1$. In such a case, the discriminative knowledge is difficult to be transferred from \emph{head} end to \emph{tail} end of the object using small dilation rates. We believe some techniques such as adversarial erasing proposed in~\cite{wei2017object} may help address this issue.

\begin{figure*}[t]
	\centering
	\includegraphics[scale=0.895]{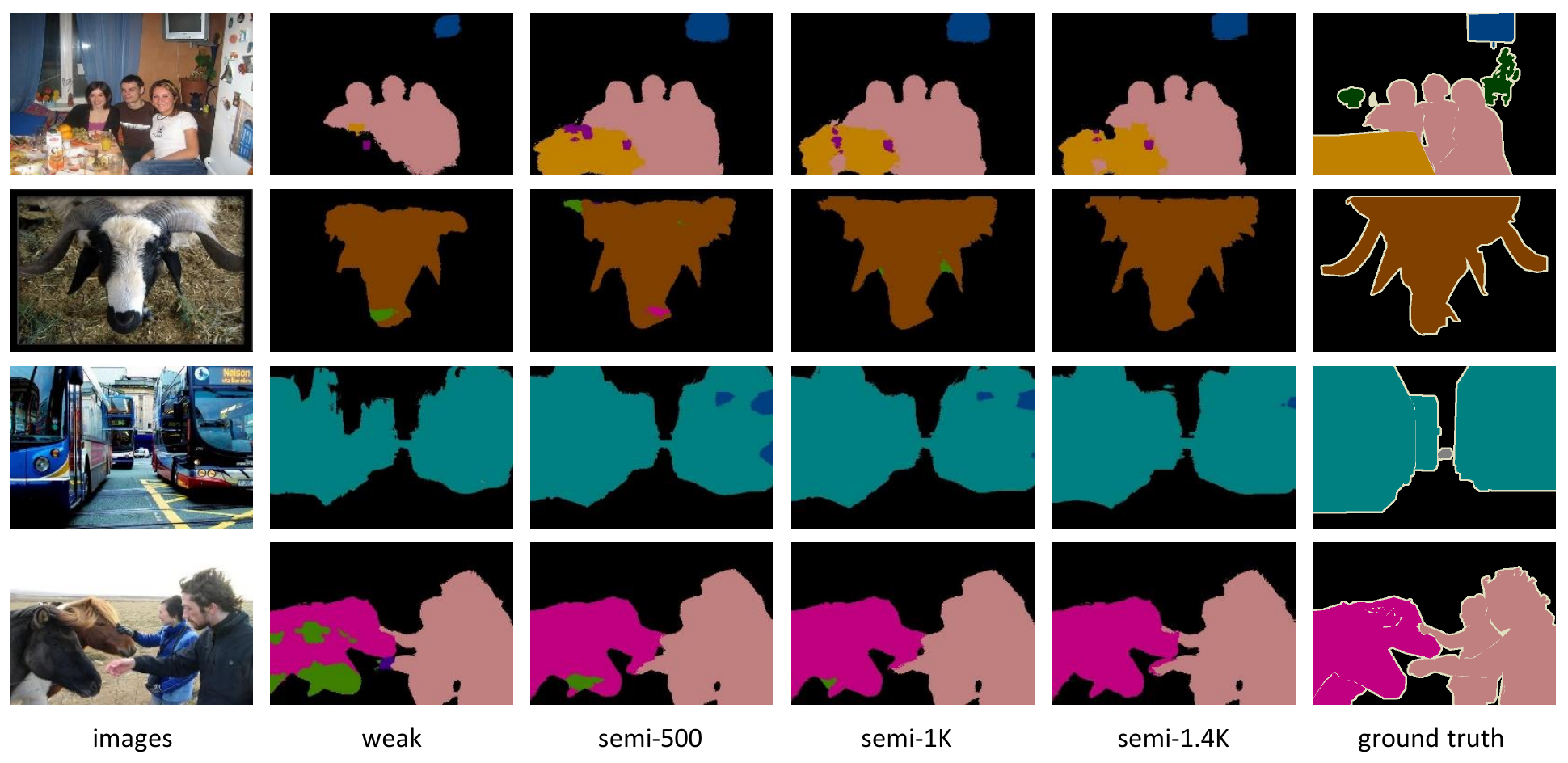}
	\caption{Examples of predicted segmentation masks by our approach in weakly- and semi- supervised manner.}
	\label{fig:samples}
	\vspace{-4mm}
\end{figure*}
\vspace{-3mm}
\subsubsection{Weakly-supervised Semantic Segmentation}

Table~\ref{tab:ablation_weak} shows the comparison of using the segmentation masks produced by different localization maps as supervision for learning segmentation networks. We observe that the performance is gradually improved (from 50.3\% to 54.4\%) by enlarging the dilation rate of the convolutional kernel, which can further validate the effectiveness of using dilated convolutional blocks for object localization. Furthermore, the mIoU score can be further improved to 57.1\% based on the dense localization maps produced by the proposed anti-noise fusion strategy, which can further demonstrate the effectiveness of this strategy for highlighting object and removing noise. Note that we also try to generate the dense localization map by averaging the localization maps from all convolutional blocks (including $d=1$). The mIoU score drops almost 1\% compared with using the current fusion strategy. Besides, there is no significant improvement in mIoU using four convolution blocks that are with the same dilation rate (\eg $d=1$) compared with that of using one block. Since conditional random field (CRF) has been considered as a standard post-processing operation for semantic segmentation and employed by all the previous works for further improving performance, we thus systematically use CRF to refine the predicted masks for a fair comparison with other state-of-the-arts. We can observe that our approach can finally achieve the mIoU score of 60.4\% and 60.8\% on validation and test sets respectively and outperform all the other weakly-supervised methods.
\vspace{-4mm}
\subsubsection{Semi-supervised Semantic Segmentation}
\vspace{-2mm}
Table~\ref{tab:ablation_semi} shows the results of using different strong/weak splits for leaning segmentation networks in semi-supervised manner. We observe that the performance only drops 0.9\% by decreasing the number of strong images from 1.4K to 500, which demonstrates that our method can easily obtain reliable segmentation results even with a small number of strong images. Based on the generated dense localization maps, we achieve new state-of-the-art results (based on 1.4K strong images) on validation and test sets with CRF post-processing. We also evaluate in another setting where using 2.9K strong images for training. We can see the corresponding mIoU score is 68.5\%, which is the same as reported in~\cite{2015-papandreou-weakly}. Since both~\cite{2015-papandreou-weakly} and this work are based on the same basic segmentation network, the performance may be saturated when the number of strongly annotated images exceeds a certain threshold. We visualize some predicted segmentation masks in Figure~\ref{fig:samples}, which shows that our approach can achieve satisfactory segmentation results w/ a few or even w/o strongly annotated images for training.

\section{Conclusion}
\vspace{-1mm}
\vspace{-0.5mm}
We revisited the dilated convolution and proposed to leverage multiple convolutional blocks of different dilated rates to generate dense object localization maps. Our approach is easy to implement and the generated dense localization maps can be utilized to learn semantic segmentation networks in weakly- or semi- supervised manner. We achieved new state-of-the-art mIoU scores on these two challenging tasks. This work paves a simple yet totally new way to mine dense object regions only with a classification network. How to address the failure cases by extending the discriminative regions from one end to the other end and conducting experiments on large-scale datasets (e.g. MS COCO~\cite{lin2014microsoft} and ImageNet~\cite{2009-imagenet}) will be our future work.



\vspace{-2mm}
\section*{Acknowledgment}
\vspace{-0.5em}
This work is supported in part by IBM-ILLINOIS Center for Cognitive Computing Systems Research (C3SR) - a research collaboration as part of the IBM AI Horizons Network, NUS startup R-263-000-C08-133, MOE Tier-I R-263-000-C21-112, NUS IDS R-263-000-C67-646 and ECRA R-263-000-C87-133. We gratefully acknowledge the support of NVIDIA Corporation and Big Data Engineering Center of E-Government, Shandong, China, with the donation of the Titan Xp GPU used for this research.

{\small
\bibliographystyle{ieee}
\bibliography{egbib}
}
\end{document}